\documentclass{article}

\usepackage{microtype}
\usepackage{graphicx}
\usepackage{subfigure}
\usepackage{booktabs} 
\usepackage{amsmath} 

\usepackage{amssymb} 
\usepackage{mathtools} 
\usepackage{amsthm} 
\usepackage{algorithm} 
\usepackage{algorithmic} 
\usepackage{multirow} 
\usepackage{siunitx} 

\usepackage{hyperref}

\usepackage[accepted]{icml2025} 

\theoremstyle{plain}

\theoremstyle{definition}

\usepackage[capitalize,noabbrev]{cleveref}

\newcommand{\methodname}{MetaGMT} 
\newcommand{\gmtlin}{GMT-LIN}
\newcommand{\gmtsam}{GMT-SAM}
\newcommand{\metagmtlin}{MetaGMT-LIN}
\newcommand{\metagmtsam}{MetaGMT-SAM}

\usepackage[textsize=tiny]{todonotes} 

\icmltitlerunning{MetaGMT: Meta-Filtered Graph Multilinear Networks} 

\begin{document}

\twocolumn[
\icmltitle{\methodname{}: Improving Actionable Interpretability of Graph Multilinear Networks via Meta-Learning Filtration}



\icmlsetsymbol{equal}{*} 

\begin{icmlauthorlist}
\icmlauthor{Rishabh Bhattacharya}{yyyy}
\icmlauthor{Hari Shankar}{yyyy}
\icmlauthor{Vaishnavi Shivkumar}{yyyy}
\icmlauthor{Ponnurangam Kumaraguru}{yyyy}
\end{icmlauthorlist}

\icmlaffiliation{yyyy}{IIIT-Hyderabad, Hyderabad, India} 

\icmlcorrespondingauthor{Rishabh Bhattacharya}{rishabh.bhattacharya@research.iiit.ac.in}
\icmlkeywords{Graph Neural Networks, Interpretability, Explainable AI, Meta-Learning, Actionable Interpretability, Subgraph Explanations}

\vskip 0.3in
]



\printAffiliationsAndNotice{} 

\begin{abstract}
The growing adoption of Graph Neural Networks (GNNs) in high-stakes domains like healthcare and finance demands reliable explanations of their decision-making processes. While inherently interpretable GNN architectures like Graph Multilinear Networks (GMT) have emerged, they remain vulnerable to generating explanations based on spurious correlations, potentially undermining trust in critical applications. We present \methodname{}, a meta-learning framework that enhances explanation fidelity through a novel bi-level optimization approach. We demonstrate that \methodname{} significantly improves both explanation quality (AUC-ROC, Precision@K) and robustness to spurious patterns, across BA-2Motifs, MUTAG, and SP-Motif benchmarks.  Our approach maintains competitive classification accuracy while producing more faithful explanations (with an increase up to $8\%$ of Explanation ROC on SP-Motif $0.5$) compared to baseline methods. These advancements in interpretability could enable safer deployment of GNNs in sensitive domains by (1) facilitating model debugging through more reliable explanations, (2) supporting targeted retraining when biases are identified, and (3) enabling meaningful human oversight. By addressing the critical challenge of explanation reliability, our work contributes to building more trustworthy and actionable GNN systems for real-world applications.
\end{abstract}  

\section{Introduction}
\label{sec:introduction}
\begin{figure*}[h]
    \centering
    \includegraphics[width=0.9\textwidth]{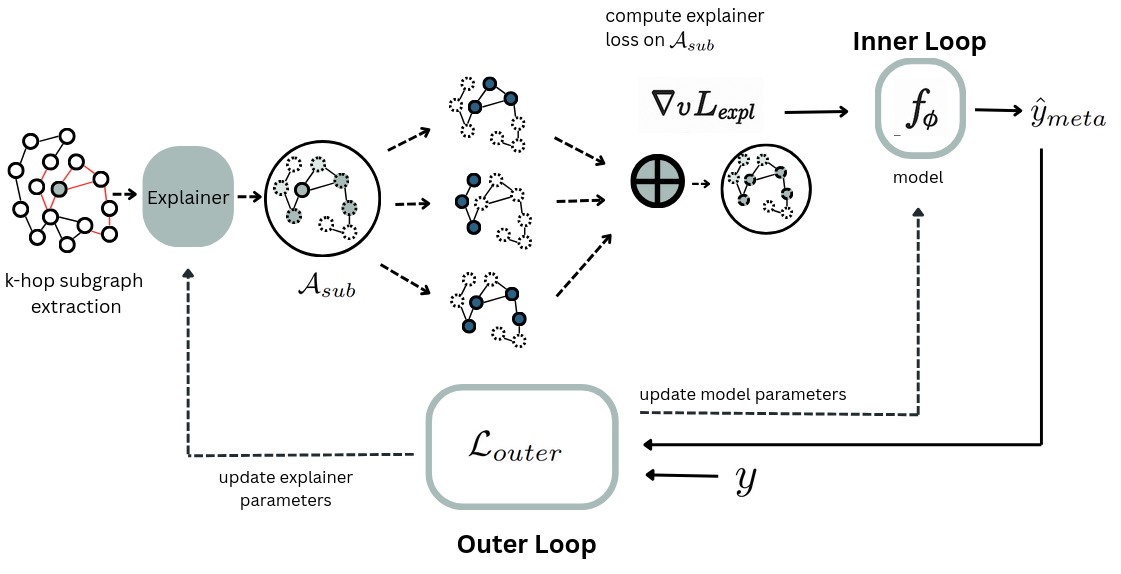}
    \caption{The complete pipeline of MetaGMT. We select a node in the graph and extract its k-hop neighbourhood. Then, we use the explainer to find the attention scores. Using the explanation loss defined in \ref{eq:info_loss} and \ref{eq:expl_loss}, We iteratively train our model (Inner Loop) to predict the correct class. This prediction is compared to ground truth (Outer Loop), and the loss calculated is used to update our explainer and our model.}
    \label{fig:metagmt-explainer}
\end{figure*}
Graph Neural Networks (GNNs) have achieved remarkable success in various domains, including bioinformatics \cite{gaudelet2021utilising}, finance \cite{yang2022graph}, and social network analysis \cite{fan2019graph}. However, their complex, non-linear nature often renders them to be black boxes, hindering trust and adoption, particularly in critical applications where understanding the rationale behind a prediction is paramount \cite{dwivedi2023explainable}. This has led to significant research into GNN interpretability, aiming to provide insights into their decision-making processes. 

We seek to obtain GNN `explanations', which may refer to subgraphs, node features or relational patterns that are critical to the input graph in order to obtain a given prediction. Existing approaches broadly fall into two categories: post-hoc methods that explain pre-trained models \cite{ying2019gnnexplainer, luo2020parameterized}, and inherently interpretable models that generate explanations concurrently with predictions \cite{miao2022interpretable, chen2024interpretable}. The latter category often employs attention mechanisms or learns explicit subgraph masks to highlight the most influential parts of the input graph.

Recently, \citet{chen2024interpretable} proposed the Graph Multilinear neT (GMT) framework, offering a principled approach to interpretable subgraph learning based on the Subgraph Multilinear Extension (SubMT). GMT, along with its variants \gmtlin{} (linear approximation) and \gmtsam{} (sampling-based approximation), represents the state-of-the-art in generating subgraph-based explanations. However, as with many explanation methods, GMT can sometimes highlight subgraphs that are strongly correlated with the prediction outcome but may not align with the underlying decision-making process of the GNN.
These correlation-based explanations can affect the interpretability fidelity, potentially leading to misinformed debugging or misguided interventions.

In order to reduce the generation of spurious explanations and consequently enhance the reliability of GNN explanations, we introduce \textbf{\methodname{}: Meta-Filtered Graph Multilinear Networks}. Our core idea is to augment the GMT training process with a meta-learning objective inspired by MATE \cite{spinelli2021meta}. 


Meta-learning, or learning-to-learn \cite{thrun1998learning}, trains models on a distribution of tasks to enable rapid adaptation to new tasks. With respect to graph explainability, the meta-objective acts as a filter, steering the optimizer and implicitly guiding the extractor to learn subgraph explanations that are sparse and predictive. Explanations based on spurious correlations are less likely to facilitate successful adaptation and will thus be penalized by the filter.



We propose and implement MetaGMT, a novel framework enhancing GMT by incorporating a meta-learning filtration step to improve explanation reliability. We demonstrate empirically on different benchmarks (BA-2Motifs, MUTAG, SP-Motif) that MetaGMT improves explanation fidelity (AUC-ROC, Precision@K) while reducing variance across random seeds, resulting in more reliable, actionable graph explanations. Figure \ref{fig:metagmt-explainer} summarizes our method. 
\section{Related Work}
\label{sec:related_work}


\subsection{GNN Interpretability}
GNN explanation methods aim to identify the crucial input components (nodes, edges, features) driving a model's prediction. We examine the two broad types of methods detailed in current literature:

\textbf{Post-hoc methods} seek to provide explanations for trained GNNs. GNNExplainer \cite{ying2019gnnexplainer} optimizes for a subgraph mask maximizing mutual information with the prediction. PGExplainer \cite{luo2020parameterized} learns a generative model for edge masks, enabling faster explanations. Other approaches include gradient-based methods \cite{pope2019explainability}, perturbation-based methods \cite{yuan2021explainability}, and surrogate models \cite{vu2020pgm}. However, while post-hoc methods are flexible,  they may not perfectly reflect the original model's internal reasoning \cite{agarwal2022evaluating}.

\textbf{Inherently interpretable methods} integrate explanation generation into the model architecture. Attention mechanisms \cite{velivckovic2017graph, miao2022interpretable} naturally provide importance scores, often regularized for sparsity. GSAT \cite{miao2022interpretable} uses stochastic gates and information bottleneck principles. SUNNY-GNN \cite{deng2023self} aims for sufficient and necessary explanations via contrastive learning. Recently, \citet{chen2024interpretable} introduced the Subgraph Multilinear Extension (SubMT) as a theoretical foundation and proposed GMT (\gmtlin{} and \gmtsam{}), which directly approximates SubMT to learn subgraph distributions, achieving state-of-the-art explanation fidelity. Our work directly builds upon and aims to enhance the GMT framework.

\subsection{Meta-Learning for Explainability and Graphs}

In the context of graph explainability, \citet{spinelli2021meta} proposed MATE (Meta-learning Approach for Training Explainable GNNs), which employs a bi-level optimization framework. The inner loop adapts a GNN using explanations from a post-hoc method (such as GNNExplainer) for sampled nodes, while the outer loop updates the GNN to learn representations that are easier to explain. Although MATE enhances a standard GNN’s explainability, it does so externally to the model’s predictive process.

In contrast, our \methodname{} adopts a similar bi-level structure, but applies it to filter explanations produced by an inherently interpretable model (GMT), thereby directly improving the relevance and fidelity of the learned explanations.

Notably, MATE’s reliance on post-hoc explainers presents limitations. Since these methods operate independently of the model’s core reasoning, their explanations are not inherently faithful and may vary significantly across random seeds or input perturbations. In comparison, \methodname{} directly refines the model’s internal explanation mechanism through meta-learning, encouraging alignment with the decision logic. By penalizing correlations that do not contribute meaningfully to the prediction, MetaGMT produces explanations that are more robust, faithful, and tightly coupled with the model’s reasoning.

\section{Preliminaries: Graph Multilinear Networks}
\label{sec:preliminaries}

Graph Multilinear neT (GMT) \cite{chen2024interpretable} aims to learn an interpretable subgraph distribution for a given input graph $G=(V, E)$ with node features $X$ and potentially edge features $E_{attr}$.

\subsection{Subgraph Multilinear Extension (SubMT)}
GMT is motivated by the Subgraph Multilinear Extension (SubMT). Let $S \subseteq E$ be a subgraph defined by a subset of edges. Let $f(S)$ be a score associated with subgraph $S$ (e.g., the GNN's prediction probability given only edges in $S$). The multilinear extension $F: [0, 1]^{|E|} \to \mathbb{R}$ of $f$ is defined for a vector $p \in [0, 1]^{|E|}$ (where $p_e$ is the probability of including edge $e$) as:
\begin{equation}
    F(p) = \sum_{S \subseteq E} f(S) \prod_{e \in S} p_e \prod_{e' \in E \setminus S} (1 - p_{e'})
\end{equation}
This represents the expected score over subgraphs sampled according to the edge probabilities $p$. GMT aims to learn edge probabilities $p$ (represented by attention scores) such that the resulting SubMT value $F(p)$ is optimized for the downstream task, while also encouraging sparsity in $p$ for interpretability.
\subsection{GMT Framework}
GMT employs two main components:

\begin{enumerate}
    \item \textbf{Classifier GNN ($\mathcal{M}_{clf}$):} A standard GNN such as a GIN or GCN \cite{xu2018powerful,kipf2016semi} mapping $(G, X) \rightarrow Y$, optionally with edge attention weights.
    \item \textbf{Extractor ($\mathcal{M}_{ext}$):} An MLP producing edge logits $\mathcal{A}_{log} \in \mathbb{R}^{|E|}$ from node embeddings, with $\mathcal{A} = \sigma(\mathcal{A}_{log})$.
\end{enumerate}

The training objective combines a prediction loss and an information loss (promoting sparsity):
\begin{equation} \label{eq:gmt_loss}
    \mathcal{L}_{GMT} = \mathcal{L}_{pred}(\mathcal{M}_{clf}(G, \mathcal{A}), y) + \lambda \mathcal{L}_{info}(\mathcal{A})
\end{equation}
where $y$ is the ground truth label, $\mathcal{L}_{pred}$ is typically cross-entropy, $\lambda$ is a balancing coefficient, and $\mathcal{L}_{info}$ is an information-theoretic regularizer (e.g., KL divergence between the attention distribution $\mathcal{A}$ and a target sparsity prior $r$):
\begin{equation} \label{eq:info_loss}
\mathcal{L}_{info}(\mathcal{A}) = \mathbb{E}_{e \sim \mathcal{A}} \left[ \log \frac{a_e}{r} + \log \frac{1-a_e}{1-r} \right]
\end{equation}
The target sparsity $r$ can be annealed during training.

\subsection{GMT Variants}
\citet{chen2024interpretable} propose two variants for integrating attention $\mathcal{A}$ into the classifier:

\begin{itemize}
    \item \textbf{\gmtlin{}:} Directly applies continuous attention scores $\mathcal{A}$ within GNN layers, linearly approximating SubMT with a single forward pass.
    
    \item \textbf{\gmtsam{}:} Draws multiple binary masks $\mathcal{m}^{(i)} \sim \text{Bernoulli}(\mathcal{A})$ and averages predictions: $\hat{y} = \frac{1}{K} \sum_{i=1}^K \mathcal{M}_{clf}(G, \mathcal{m}^{(i)})$. This yields a better SubMT approximation but incurs higher computational cost.
\end{itemize}

\section{\methodname{}: Meta-Filtered GMT}
\label{sec:method}

While GMT provides a strong baseline for interpretable subgraph learning, the learned attention masks $\mathcal{A}$ can still highlight edges that are spuriously correlated with the label rather than being truly explanatory. This limits the faithfulness and reliability of the resulting explanations.

To address this, we propose \methodname{}, which enhances GMT with a meta-learning-based filtration strategy. The key idea is to encourage explanations that are not just predictive within a local subgraph but also generalizable to the full graph. Rather than relying solely on static loss functions, \methodname{} evaluates the utility of explanations dynamically—penalizing those that fail to support consistent model behavior across contexts. The following subsections formalize this idea. We summarize our pipeline in Figure \ref{fig:metagmt-explainer}.

\subsection{Rationale: Filtration via Meta-Adaptation}


Let \( G = (V, E) \) be an input graph, and \( G_{\text{sub}} \subset G \) be a \(k\)-hop subgraph centered at node \(v \in V\). Given attention scores \( \mathcal{A}_{\text{sub}} \) produced by the extractor \( \mathcal{M}_{ext} \), a local explanation is defined by a sparse mask over \( E_{\text{sub}} \). Intuitively, if this explanation captures the true decision-relevant structure, then adapting the classifier \( \mathcal{M}_{clf} \) using \( \mathcal{A}_{\text{sub}} \) should improve performance on the original graph \( G \).

Conversely, explanations relying on spurious correlations will degrade full-graph performance when used for adaptation. \methodname{} exploits this principle by filtering explanations through a bi-level optimization loop, described in the next section, where the outer loss on \( G \) provides feedback on the quality of explanations used in the inner loop.

\subsection{Bi-Level Optimization}

\begin{algorithm}[t]
\caption{MetaGMT Training Step}
\label{alg:metagmt}
\begin{algorithmic}[1]
   \STATE {\bfseries Input:}
   \begin{itemize}
      \item Graph $G = (X, V, E)$, label $y$
      \item Classifier $\mathcal{M}_{clf}(\theta_{clf})$, extractor $\mathcal{M}_{ext}(\theta_{ext})$
      \item Inner learning rate $\alpha$, outer optimizer $Opt_{outer}$
      \item Inner steps $N_{inner}$, hops $k$
   \end{itemize}
   
   \STATE Sample node $v \in V$.
   \STATE Extract $k$-hop subgraph $G_{sub}$ around $v$.
   \STATE Compute subgraph attention $\mathcal{A}_{sub}$ using $\mathcal{M}_{ext}(\theta_{ext})$.
   \STATE Initialize functional classifier parameters $\phi_{clf}^{(0)} \leftarrow \theta_{clf}$.
   \STATE Initialize inner optimizer $Opt_{inner}$ (e.g., SGD) for $\phi_{clf}$.
   
   \STATE {\bfseries Inner Loop:}
   \FOR{$n=0$ {\bfseries to} $N_{inner}-1$}
       \STATE Compute $\mathcal{L}_{expl}(\mathcal{A}_{sub})$ (Eq. \ref{eq:expl_loss}).
       \STATE Update $\phi_{clf}^{(n+1)} \leftarrow Opt_{inner}(\phi_{clf}^{(n)}, \nabla_{\phi_{clf}^{(n)}} \mathcal{L}_{expl})$.
   \ENDFOR
   \STATE Let $\phi_{clf}^{*} = \phi_{clf}^{(N_{inner})}$.
   
   \STATE {\bfseries Outer Loop:}
   \STATE Compute meta-prediction $\hat{y}_{meta} = \mathcal{M}_{clf}(\phi_{clf}^{*})(G)$. 
   \STATE Compute outer loss $\mathcal{L}_{outer} = \mathcal{L}_{pred}(\hat{y}_{meta}, y)$.
   \STATE Compute meta-gradients:
   \STATE \hspace{0.5cm} $g_{clf} = \nabla_{\theta_{clf}} \mathcal{L}_{outer}$
   \STATE \hspace{0.5cm} $g_{ext} = \nabla_{\theta_{ext}} \mathcal{L}_{outer}$.
   \STATE Update $\theta_{clf} \leftarrow Opt_{outer}(\theta_{clf}, g_{clf})$.
   \STATE Update $\theta_{ext} \leftarrow Opt_{outer}(\theta_{ext}, g_{ext})$.
\end{algorithmic}
\end{algorithm}

We train the model architectures from scratch. Let $\theta_{clf}$ be the parameters of the classifier $\mathcal{M}_{clf}$ and $\theta_{ext}$ be the parameters of the extractor $\mathcal{M}_{ext}$. The standard GMT training updates both $\theta_{clf}$ and $\theta_{ext}$ based on $\mathcal{L}_{GMT}$ (Eq. \ref{eq:gmt_loss}). \methodname{} modifies this update as follows for each training batch (or graph) $G$:



\textbf{1. Inner Loop: Classifier Adaptation on Subgraph}

The process begins by sampling a node \( v \in V \) uniformly at random. From this node, the \( k \)-hop computational subgraph \( G_{sub} = (V_{sub}, E_{sub}) \) centered around \( v \) is extracted, where \( X_{sub} \) and \( E_{attr, sub} \) denote the corresponding node and edge features, respectively (in our implementation, \( k=1 \)). The original extractor \( \mathcal{M}_{ext}(\theta_{ext}) \) then processes the initial node embeddings of the subgraph to compute attention logits \( \mathcal{A}_{log, sub} \) and normalized attention scores \( \mathcal{A}_{sub} = \sigma(\mathcal{A}_{log, sub}) \) for the edges \( E_{sub} \), based on the initial node representations \( \mathcal{h}_{sub}^{(0)} \).  

An explanation loss \( \mathcal{L}_{expl} \) is defined on the subgraph, specifically using the information loss (Eq. \ref{eq:info_loss}) applied to the subgraph attention scores:

\begin{equation} \label{eq:expl_loss}  
    \mathcal{L}_{expl}(\mathcal{A}_{sub}) = \mathcal{L}_{info}(\mathcal{A}_{sub})  
\end{equation}  

This loss encourages the extractor to produce sparse and focused attention patterns, even within the local subgraph context.  

Next, a functional copy of the classifier \( \mathcal{M}_{clf} \) is created with parameters \( \phi_{clf} \), initialized to the original classifier parameters \( \theta_{clf} \). The inner loop adaptation consists of \( N_{inner} \) steps, where \( \phi_{clf} \) is updated using an inner optimizer (e.g., SGD with learning rate \( \alpha \)) to minimize \( \mathcal{L}_{expl} \):  

\begin{equation}  
    \phi_{clf}^{(n+1)} = \phi_{clf}^{(n)} - \alpha \nabla_{\phi_{clf}^{(n)}} \mathcal{L}_{expl}(\mathcal{A}_{sub})  
\end{equation}  

The final adapted parameters after \( N_{inner} \) steps are denoted by \( \phi_{clf}^{*} = \phi_{clf}^{(N_{inner})} \). The inner-loop optimization updates \( \phi_{clf} \), while the extractor parameters \( \theta_{ext} \) remain fixed throughout the adaptation process.

\textbf{2. Outer Loop: Meta-Update on Full Graph}

We apply the adapted classifier \(\mathcal{M}_{clf}(\phi_{clf}^{*})\) to the original full graph \(G\), yielding the prediction \(\hat{y}_{meta} = \mathcal{M}_{clf}(\phi_{clf}^{*})(G)\). The adapted classifier retains the original extractor \(\mathcal{M}_{ext}(\theta_{ext})\) to generate attention weights \(\mathcal{A}\) for the full graph during its forward pass, consistent with the standard GMT forward pass.  

Next, we compute the standard classification loss on the full graph using the adapted classifier's prediction:  

\begin{equation} \label{eq:outer_loss}  
    \mathcal{L}_{outer} = \mathcal{L}_{pred}(\hat{y}_{meta}, y)  
\end{equation}  

The gradients of \(\mathcal{L}_{outer}\) with respect to the original parameters \(\theta_{clf}\) and \(\theta_{ext}\) are then calculated. The `higher' library facilitates this by automatically backpropagating through the inner loop adaptation steps, producing:  

\begin{equation}  
    g_{clf} = \nabla_{\theta_{clf}} \mathcal{L}_{outer} \quad ; \quad g_{ext} = \nabla_{\theta_{ext}} \mathcal{L}_{outer}  
\end{equation}  

Finally, the original parameters are updated using the main optimizer, such as Adam with learning rate \(\beta\):  

\begin{equation}  
    \theta_{clf} \leftarrow \text{OptimizerUpdate}(\theta_{clf}, g_{clf})  
\end{equation}  
\begin{equation}  
    \theta_{ext} \leftarrow \text{OptimizerUpdate}(\theta_{ext}, g_{ext})  
\end{equation}  

The overall process is summarized in Algorithm \ref{alg:metagmt}.

\begin{table*}[t]
\caption{Interpretation Performance (ROC) across BA-2Motifs, MUTAG and Spurious-Motif benchmarks. Methods having better performance are highlighted in bold. Results highlight the robustness of MetaGMT-LIN compared to baseline GMT-LIN, particularly in high-bias settings (SP-Motif, b = 0.9), where leveraging meta-learning improves invariance to spurious features. Performance on BA-2Motifs and MUTAG (alternative benchmarks) is included for reference, showing consistent competitiveness. Standard deviations reflect stability across different seeds (0-9 inclusive).}
\label{tab:roc_results}
\begin{center}
\begin{small}
\begin{sc}
\sisetup{round-mode=places, round-precision=2, table-format=2.2(2)}
\begin{tabular}{llllll}
\hline
            & \textbf{BA-2Motifs} & \textbf{Mutag}   & \multicolumn{3}{c}{\textbf{Spurious-Motif}}            \\ \cline{4-6} 
            &                     &                  & \textbf{b = 0.5} & \textbf{b = 0.7} & \textbf{b = 0.9} \\ \hline
GMT-LIN     & 97.97 $\pm$1.46     & \textbf{93.13 $\pm$ 2.06} & 72.62 $\pm$ 7.99   & 72.14 $\pm$ 14.89  & 66.84 $\pm$ 12.36  \\
MetaGMT-LIN & \textbf{98.37 $\pm$1.08}     & 92.60 $\pm$ 2.05   & \textbf{80.04 $\pm$ 2.76}   & \textbf{80.16 $\pm$ 8.28}   &   \textbf{69.59 $\pm$ 4.18}                \\ \hline
\end{tabular}
\end{sc}
\end{small}
\end{center}
\end{table*}

\begin{table*}[t]
\caption{Explanation Precision @ 5 (X-Prec@5) across BA-2Motifs, MUTAG and Spurious-Motif benchmarks. Methods having better performance are highlighted in bold. MetaGMT-LIN demonstrates superior robustness at b = 0.9, outperforming baselines by $9\%$ (50.79 vs. 41.85), suggesting better avoidance of spurious explanations. On BA-2Motifs, MetaGMT-LIN achieves high precision ($>$ 88\%), while MUTAG (real-world) results reflect inherent task difficulty. Standard deviations highlight variance across seeds.}
\label{tab:xprec_results}
\begin{center}
\begin{small}
\begin{sc}
\sisetup{round-mode=places, round-precision=2, table-format=2.2(2)}
\begin{tabular}{llllll}
\hline
            & \textbf{BA-2Motifs} & \textbf{Mutag} & \multicolumn{3}{c}{\textbf{Spurious-Motif}}            \\ \cline{4-6} 
            &                     &                & \textbf{b = 0.5} & \textbf{b = 0.7} & \textbf{b = 0.9} \\ \hline
GMT-LIN     & 88.58 $\pm$10.60        & \textbf{28.66 $\pm$ 1.38} & 46.34 $\pm$ 6.43   & 48.66 $\pm$ 13.13  & 41.85 $\pm$ 9.29   \\
MetaGMT-LIN & \textbf{91.96 $\pm$3.79}         & 28.55 $\pm$ 1.92 & \textbf{48.48 $\pm$ 2.96}   & \textbf{57.82 $\pm$ 9.71}   &  \textbf{50.79 $\pm$ 6.12}               \\ \hline
\end{tabular}
\end{sc}
\end{small}
\end{center}
\end{table*}

\section{Experiments}
\label{sec:experiments}

\subsection{Experimental Setup}

\textbf{Dataset:} We use the BA-2Motifs \cite{ying2019gnnexplainer}, MUTAG \cite{zhang2019hierarchical} and Sp-Motif \cite{wanglearning} benchmarks to evaluate our methodology. BA-2Motifs is a synthetic benchmark commonly used for evaluating GNN explanation methods. It consists of Barabási-Albert (BA) graphs where specific ``house" or ``cycle" motifs are attached. The task is graph classification based on the presence of the motif. Ground truth edge explanations correspond to the edges forming the attached motif.

MUTAG is a real-world benchmark dataset for graph classification, consisting of 188 molecular graphs representing nitroaromatic compounds. Each graph corresponds to a chemical compound where nodes represent atoms and edges represent chemical bonds. The primary classification task is to predict the mutagenicity of these compounds, i.e., their potential to cause mutations in the DNA of Salmonella typhimurium. Ground truth explanations often correspond to specific chemical substructures such as carbon rings containing $NH_2$ or $NO_2$ groups that are known to be mutagenic.

SP-Motif is a synthetic benchmark designed to evaluate graph learning methods under distribution shifts and spurious correlations. Each graph in SP-Motif consists of two parts: a variant base subgraph and an invariant motif subgraph. The graph label depends solely on the motif subgraph, which represents the true causal structure, while the base subgraph is spuriously correlated with the label during training. SP-Motif additionally provides a parameter $b$ that controls the degree of spurious correlation between the non-causal base subgraph and the graph label during training. The parameter $b$ determines the probability that a training sample's base structure (e.g., a tree, wheel, or ladder) aligns with the label, even though the label is causally determined only by the motif. 

We present our results for \gmtlin{} on SP-Motif with three different configurations: $b = 0.5$, $b = 0.7$ and $b = 0.9$. For \gmtsam{} we present our results for $b = 0.5$, and include runs for $b=0.7$ and $b=0.9$ in the GitHub repository\footnote{Code available at \href{https://anonymous.4open.science/r/goddammit-sparky-F9DB/}{https://anonymous.4open.science/r/goddammit-sparky-F9DB/}}.  

\textbf{Baselines:} We compare our meta-filtered variants, \metagmtlin{} and \metagmtsam{}, against their respective vanilla methods, \gmtlin{} and \gmtsam{} \cite{chen2024interpretable}, trained using the standard objective (Eq. \ref{eq:gmt_loss}).

\textbf{Implementation:} We adapt the official GMT implementation based on PyTorch Geometric \cite{fey2019fast}. The backbone GNN classifier uses a GIN architecture. The extractor is an MLP. Key meta-parameters were set as follows: inner steps $N_{inner}=3$, inner learning rate $\alpha=0.01$, subgraph hops $k=1$. Other hyperparameters (outer learning rate, loss coefficients, sparsity schedule, etc.) were kept consistent between the baseline GMT and \methodname{} variants, following the optimal settings reported for GMT on datasets where applicable, or tuned slightly on a validation set. We report results averaged over 10 different seeds, from 0 to 9.

\textbf{Evaluation Metrics:} We evaluate both explanation quality and classification performance:
\begin{itemize}
    \item \textbf{Explanation AUC (X-ROC):} Area Under the ROC Curve comparing the learned edge attention scores against the ground truth motif edge labels. Measures explanation fidelity.
    \item \textbf{Explanation Precision@K (X-Prec@K):} Precision of the top-K predicted edges based on attention scores, compared to ground truth motif edges. We use K=5, matching the typical motif size. Measures explanation accuracy for the most salient edges.
    \item \textbf{Classification Accuracy (Clf-Acc):} Standard graph classification accuracy on the test set.
\end{itemize}
We report the metrics achieved at the epoch yielding the best validation classification accuracy.

\subsection{Results}
Tables \ref{tab:roc_results} and \ref{tab:xprec_results} present comprehensive performance comparisons between \gmtlin{} and our proposed \metagmtlin{} variants across BA-2Motifs, MUTAG, and Spurious-Motif benchmarks.

Our results demonstrate consistent improvements in both interpretation quality and model robustness. On BA-2Motifs, \metagmtlin{} achieves superior interpretation performance with 98.37\% X-ROC (compared to 97.97\% for \gmtlin{}) and 91.96\% X-Prec@5 (vs 88.58\%). 

The stability of explanations shows notable enhancement across all benchmarks. For BA-2Motifs, the standard deviation of X-Prec@5 decreases from $\pm$10.60 for \gmtlin{} to just $\pm$3.79 for \metagmtlin{}, indicating significantly more consistent explanation quality. This pattern holds for the Spurious-Motif benchmark, where \metagmtlin{} achieves both higher mean scores and lower variance compared to baseline methods. 


The performance improvements are particularly pronounced in challenging scenarios with higher bias levels in the Spurious-Motif benchmark. At bias level 0.7, \metagmtlin{} achieves 80.16\% X-ROC ($\pm$8.28) compared to 72.14\% ($\pm$14.89) for \gmtlin{}, demonstrating both higher performance and greater stability in the presence of confounding factors. All metrics have been reported in Tables \ref{tab:roc_results} and \ref{tab:xprec_results}.

Tables \ref{tab:roc_sam} and \ref{tab:xprec_sam} present performance comparisons between \gmtsam{} and the proposed \metagmtsam{}. Overall, MetaGMT-SAM shows slight improvements over GMT-SAM in ROC scores on BA-2Motifs (97.83 vs. 97.13) and Spurious-Motif (79.25 vs. 73.78), indicating enhanced interpretability. However, it performs marginally worse on MUTAG (93.58 vs. 94.79). In terms of X-Prec@5, GMT-SAM slightly outperforms MetaGMT-SAM on BA-2Motifs and Spurious-Motif (b = 0.5), while both models yield nearly identical results on MUTAG. These results suggest that Meta-learning benefits ROC-based interpretation performance more consistently than precision-based explanation accuracy.

\begin{table}[h]
\caption{ROC-AUC scores comparing GMT-SAM and MetaGMT-SAM across benchmarks.Methods having better performance are highlighted in bold.  MetaGMT-SAM shows consistent improvements on synthetic datasets, achieving +0.7\% on BA-2Motifs (97.83 vs 97.13) and a more substantial +5.47\% gain on SP-Motif with b=0.5 (79.25 vs 73.78), demonstrating better robustness to spurious correlations.}
\label{tab:roc_sam}
\begin{center}
\begin{small}
\begin{sc}
\sisetup{round-mode=places, round-precision=2, table-format=2.2(2)}
\resizebox{\columnwidth}{!}{
\begin{tabular}{lccc}
\hline
            & \textbf{BA-2Motifs} & \textbf{Mutag} & \textbf{SP-Motif (b = 0.5)} \\ \hline
GMT-SAM     & 97.13 $\pm$ 1.28 & \textbf{94.79 $\pm$ 2.07} & 73.78 $\pm$ 8.82 \\
MetaGMT-SAM & \textbf{97.83 $\pm$ 1.65} & 93.58 $\pm$ 3.10 & \textbf{79.25 $\pm$ 3.07} \\ \hline
\end{tabular}
}
\end{sc}
\end{small}
\end{center}
\end{table}

\begin{table}[h]
\caption{Explanation Precision@5 comparison for GMT-SAM and MetaGMT-SAM. Methods having better performance are highlighted in bold. MetaGMT-SAM achieves higher precision on BA-2Motifs (91.72 vs 90.36) with lower variance ($\pm$2.70 vs $\pm$3.77), suggesting more reliable explanations for synthetic graphs. On SP-Motif (b=0.5), GMT-SAM shows marginally better performance (50.14 vs 48.72) but with nearly double the standard deviation (7.83 vs 3.63), making MetaGMT-SAM's explanations more consistent.}
\label{tab:xprec_sam}
\begin{center}
\begin{small}
\begin{sc}
\sisetup{round-mode=places, round-precision=2, table-format=2.2(2)}
\resizebox{\columnwidth}{!}{
\begin{tabular}{lccc}
\hline
            & \textbf{BA-2Motifs} & \textbf{Mutag} & \textbf{SP-Motif (b = 0.5)} \\ \hline
GMT-SAM     & 90.36 $\pm$ 3.77 & 2\textbf{8.65 $\pm$ 1.43} & \textbf{50.14 $\pm$ 7.83} \\
MetaGMT-SAM & \textbf{91.72 $\pm$ 2.70} & 28.48 $\pm$ 1.92 & 48.72 $\pm$ 3.63 \\ \hline
\end{tabular}
}
\end{sc}
\end{small}
\end{center}
\end{table}

These results collectively demonstrate that our proposed methodology not only enhances explanation fidelity and classification accuracy but also significantly reduces performance variability across different seeds and data splits. Consistent improvements across multiple benchmarks and metrics suggest that the meta-learning filtration step provides superior generalization capabilities on unseen data.

\section{Discussion}

Consistent improvements in explanation fidelity metrics (X-ROC, X-Prec@5) suggest that the meta-learning filtration mechanism successfully guides the extractor to focus on edges that are more aligned with the ground truth motifs. By rewarding explanations that facilitate robust classifier adaptation from local subgraphs to the global graph, \methodname{} effectively filters out potentially spurious edges that might be captured by the standard GMT objective alone.

This enhanced reliability is crucial for actionable interpretability. We list some of the potential use cases below:

\begin{itemize}
    \item \textbf{Model Debugging:} If the model makes an error, a high-fidelity explanation is more likely to pinpoint the correct cause (e.g., a missing or distorted motif), enabling targeted corrections. In contrast, a misleading explanation might send developers in the wrong direction.
    \item \textbf{Scientific Discovery:} In domains like drug discovery or materials science, identifying the true structural drivers of a property is critical. More faithful and robust explanations improve the likelihood of discovering meaningful structure-property relationships.
    \item \textbf{Trust and Verification:} Users are more likely to trust and rely on explanations that consistently highlight relevant features across varying contexts.
    
    \item \textbf{Model Retraining and Deployment:} Reliable explanations can guide selective model editing and retraining, allowing developers to iteratively improve generalization or correct failure modes without exhaustive manual inspection.
\end{itemize}

Our experiments further show that improvements in interpretability come with either preserved or slightly improved predictive performance, indicating that the meta-learning objective helps regularize the model in a way that benefits both explanation quality and robustness. This robustness is particularly important in real-world deployments, where consistent performance across retraining runs or environments is crucial.


The meta-objective provides a self-supervised signal for explanation quality, rooted in internal consistency and adaptability. It encourages the model to favor explanations that are not only faithful but also useful, thus aligning with the goals of actionable interpretability and practical impact.

\section{Limitations and Future Work}

Incorporating a meta-learning filtration step in the standard GMT training step comes with a notable increase in computational cost, due to the bi-level optimization taking place at each epoch. Consequently, the computational cost scales with the number of inner loop adaptation steps. Our experimental setup used a fixed number of inner steps $N_{inner} = 3$ for all datasets. 

In practice, incorporating a bi-level optimization method, by means of meta-learning, we introduce a trade-off between the benefits of meta-learning for improved interpretability and the computational overhead required during training, necessitating careful consideration of resource constraints when implementing MetaGMT.




By producing more robust, reliable explanations, \methodname{} contributes to the goal of actionable interpretability for GNNs. Future work includes evaluating \methodname{} on more complex real-world datasets, exploring different formulations of the inner-loop explanation loss, and investigating the computational trade-offs of the meta-learning approach. We believe that integrating principles of adaptation and generalization, as done in \methodname{}, is a promising direction for developing GNN explanation methods that are not just interpretable, but truly trustworthy and actionable. Trustworthy explanations may be beneficial for the adaptation of such models for tasks in various domains such as e-commerce, social networks, recommendation systems, etc.  

\section*{Impact Statement}
This paper presents work whose goal is to advance the field of Machine Learning, specifically in the area of Graph Neural Network interpretability. By improving the reliability and actionability of GNN explanations, our work aims to foster greater trust and enable more responsible deployment of GNNs in critical applications such as healthcare, finance, and scientific discovery. While improved interpretability can lead to positive societal outcomes through better model understanding and debugging, potential negative consequences could arise if misinterpreted or over-relied upon explanations lead to flawed decisions. We believe the primary impact is positive, contributing to more transparent and trustworthy AI systems. There are many potential societal consequences of our work, none which we feel must be specifically highlighted here beyond the general implications of advancing interpretable machine learning.


\nocite{langley00} 

\bibliography{references} 
\bibliographystyle{icml2025}

\newpage
\appendix
\onecolumn 
\section{Appendix}

\subsection{Classification Accuracy}

\begin{table*}[h]
\caption{Classification Accuracy (Clf-Acc) across BA-2Motifs, MUTAG and Spurious-Motif benchmarks. \gmtlin{} outperforms \metagmtlin{} in high-bias scenarios (b = 0.9), with GMT-LIN achieving 66.90\% accuracy compared to \metagmtlin{}'s 54.53\%, suggesting potential trade-offs between classification performance and explanation quality in biased settings.}
\label{tab:clfacc_results}
\vskip 0.15in
\begin{center}
\begin{small}
\begin{sc}
\sisetup{round-mode=places, round-precision=2, table-format=2.2(2)}
\begin{tabular}{llllll}
\hline
            & \textbf{BA-2Motifs} & \textbf{Mutag} & \multicolumn{3}{c}{\textbf{Spurious-Motif}}            \\ \cline{4-6} 
            &                     &                & \textbf{b = 0.5} & \textbf{b = 0.7} & \textbf{b = 0.9} \\ \hline
GMT-LIN     & 99.20 $\pm$0.98         & 91.56 $\pm$ 0.85 & 75.13 $\pm$ 1.22   & 69.37 $\pm$ 1.67   & 66.90 $\pm$ 4.48   \\
MetaGMT-LIN & 99.50 $\pm$0.50         & 91.12 $\pm$ 1.05 & 71.64 $\pm$ 1.97   & 65.00 $\pm$ 3.51   & 54.53 $\pm$ 2.09                \\ \hline
\end{tabular}
\end{sc}
\end{small}
\end{center}
\end{table*}

MetaGMT-LIN achieves equal or superior accuracy to their baseline counterparts on the BA-2Motifs dataset. Notably, the method shows significantly reduced variance compared to the baselines, particularly for \gmtsam{} where the standard deviation decreases from $\pm$2.40 to just $\pm$0.66.

For the MUTAG benchmark, classification accuracies remain stable between baseline and MetaGMT variants, with all methods achieving between 90.07\% and 91.56\% accuracy. The minor differences ($\leq$ 1.5\%) suggest that our meta-learning approach maintains strong discriminative performance while improving explanation quality.

Results on the Spurious-Motif benchmark indicate a potential trade-off pattern. While MetaGMT variants show slightly lower mean accuracy (71.42\%-71.64\% vs 75.13\%-75.49\% at b=0.5), they maintain more consistent performance as bias increases, as evidenced by smaller accuracy drops across bias levels. This suggests our method may sacrifice marginal classification performance for greater robustness against spurious correlations.


\clearpage

\subsection{Hyperparameter Details}

Tables \ref{tab:common_configs} and \ref{tab:dataset_configs} summarize our experimental configurations. Following the GMT framework established by \citet{chen2024interpretable}, we maintain consistent hyperparameters where applicable while adapting specific settings (learning rates, attention mechanisms, and sparsity thresholds) to each benchmark's characteristics. 

\begin{table}[h]
\centering
\caption{Common hyperparameters across all benchmarks}
\label{tab:common_configs}
\small
\begin{tabular}{lc}
\hline
\textbf{Parameter} & \textbf{Value} \\ \hline
\textbf{Architecture} & \\
\quad Backbone & GIN (2 layers, 64 dim) \\
\quad Dropout & 0.3 \\
\quad Extractor dropout & 0.5 \\ \hline

\textbf{Training} & \\
\quad Precision@K & 5 \\
\quad Visualization interval & Every 10 epochs \\
\quad Training epochs & 100 \\
\quad $\lambda_{pred}$, $\lambda_{info}$ & 1, 1 \\
\quad Sparsity decay schedule & -0.1 every 10 epochs \\ \hline
\end{tabular}
\end{table}

\begin{table}[h]
\centering
\caption{Dataset-specific configurations}
\label{tab:dataset_configs}
\small
\begin{tabular}{lccc}
\hline
\textbf{Parameter} & \textbf{BA-2Motifs} & \textbf{MUTAG} & \textbf{SP-Motif} \\ \hline
\textbf{Data} & & & \\
\quad Split & 80/10/10 & 80/10/10 & - \\
\quad Batch size & 128 & 128 & 128 \\ \hline

\textbf{Training} & & & \\
\quad Pretrain lr & 1e-3 & 1e-3 & 3e-3 \\
\quad Learning rate & 1e-3 & 1e-3 & 3e-3 \\ \hline

\textbf{Options} & & & \\
\quad Use Edge Attention & Yes & No & Yes \\

\textbf{Sparsity} & & & \\
\quad Initial $r$ & 1.0 & 1.0 & 1.0 \\
\quad Final $r$ & 0.5 & 0.5 & 0.7 \\ \hline
\end{tabular}
\end{table}



\end{document}